\documentclass[12pt, onecolumn, draftclsnofoot]{IEEEtran}%

\usepackage[T1]{fontenc}
\usepackage{amsmath,amssymb,amsfonts,mathrsfs,bm}
\usepackage{mathtools}
\usepackage{amsthm}
\usepackage{cite}
\usepackage[shortlabels]{enumitem}
\usepackage{enumitem}
\usepackage{graphicx}
\usepackage{epstopdf}
\usepackage{url}
\usepackage{colortbl}
\usepackage{multirow}
\usepackage[table,dvipsnames]{xcolor}
\usepackage[normalem]{ulem}
\usepackage{array}
\newcolumntype{L}[1]{>{\raggedright\let\newline\\\arraybackslash\hspace{0pt}}m{#1}}
\newcolumntype{C}[1]{>{\centering\let\newline\\\arraybackslash\hspace{0pt}}m{#1}}
\newcolumntype{R}[1]{>{\raggedleft\let\newline\\\arraybackslash\hspace{0pt}}m{#1}}

\makeatletter
\let\MYcaption\@makecaption
\makeatother
\usepackage[font=footnotesize]{subcaption}
\makeatletter
\let\@makecaption\MYcaption
\makeatother

\usepackage{xparse}



\usepackage{glossaries}
\newacronym{wrt}{w.r.t.}{with respect to}
\newacronym{RHS}{R.H.S.}{right-hand side}
\newacronym{LHS}{L.H.S.}{left-hand side}
\newacronym{iid}{i.i.d.}{independent and identically distributed}

\usepackage{float}

\ifx\notloadhyperref\undefined
	\ifx\loadbibentry\undefined
		\usepackage[hidelinks,hypertexnames=false]{hyperref} 
	\else
		\usepackage{bibentry}
		\makeatletter\let\saved@bibitem\@bibitem\makeatother
		\usepackage[hidelinks,hypertexnames=false]{hyperref}
		\makeatletter\let\@bibitem\saved@bibitem\makeatother
	\fi
\else
	\relax
\fi

\usepackage[capitalize]{cleveref}
\crefname{equation}{}{}
\Crefname{equation}{}{}
\crefname{claim}{claim}{claims}
\crefname{step}{step}{steps}
\crefname{line}{line}{lines}
\crefname{condition}{condition}{conditions}
\crefname{dmath}{}{}
\crefname{dseries}{}{}
\crefname{dgroup}{}{}
\crefname{Theorem}{Theorem}{Theorems}
\crefname{Corollary}{Corollary}{Corollaries}
\crefname{Proposition}{Proposition}{Propositions}
\crefname{Lemma}{Lemma}{Lemmas}
\crefname{Definition}{Definition}{Definitions}
\crefname{Example}{Example}{Examples}
\crefname{Assumption}{Assumption}{Assumptions}
\crefname{Remark}{Remark}{Remarks}
\crefname{Rem}{Remark}{Remarks}
\crefname{remarks}{Remarks}{Remarks}
\crefname{Exercise}{Exercise}{Exercises}
\crefname{Theorem_A}{Theorem}{Theorems}
\crefname{Corollary_A}{Corollary}{Corollaries}
\crefname{Proposition_A}{Proposition}{Propositions}
\crefname{Lemma_A}{Lemma}{Lemmas}
\crefname{Definition_A}{Definition}{Definitions}

\usepackage{algorithm,algorithmic}

\ifx\loadbreqn\undefined
	\relax
\else
	\usepackage{breqn} 
\fi

\ifx\renewtheorem\undefined
\ifx\useTheoremCounter\undefined
\newtheorem{Theorem}{Theorem}
\newtheorem{Corollary}{Corollary}
\newtheorem{Proposition}{Proposition}
\newtheorem{Lemma}{Lemma}
\else
\newtheorem{Theorem}{Theorem}
\newtheorem{Corollary}[theorem]{Corollary}
\newtheorem{Proposition}[theorem]{Proposition}
\fi

\newtheorem{Definition}{Definition}
\newtheorem{Example}{Example}

\newtheorem{Lemma_A}{Lemma}[section]

\fi

\theoremstyle{remark}

\theoremstyle{plain}

\DeclareSymbolFont{bsfletters}{OT1}{cmss}{bx}{n}
\DeclareSymbolFont{ssfletters}{OT1}{cmss}{m}{n}
\DeclareMathSymbol{\bsfGamma}{0}{bsfletters}{'000}
\DeclareMathSymbol{\ssfGamma}{0}{ssfletters}{'000}
\DeclareMathSymbol{\bsfDelta}{0}{bsfletters}{'001}
\DeclareMathSymbol{\ssfDelta}{0}{ssfletters}{'001}
\DeclareMathSymbol{\bsfTheta}{0}{bsfletters}{'002}
\DeclareMathSymbol{\ssfTheta}{0}{ssfletters}{'002}
\DeclareMathSymbol{\bsfLambda}{0}{bsfletters}{'003}
\DeclareMathSymbol{\ssfLambda}{0}{ssfletters}{'003}
\DeclareMathSymbol{\bsfXi}{0}{bsfletters}{'004}
\DeclareMathSymbol{\ssfXi}{0}{ssfletters}{'004}
\DeclareMathSymbol{\bsfPi}{0}{bsfletters}{'005}
\DeclareMathSymbol{\ssfPi}{0}{ssfletters}{'005}
\DeclareMathSymbol{\bsfSigma}{0}{bsfletters}{'006}
\DeclareMathSymbol{\ssfSigma}{0}{ssfletters}{'006}
\DeclareMathSymbol{\bsfUpsilon}{0}{bsfletters}{'007}
\DeclareMathSymbol{\ssfUpsilon}{0}{ssfletters}{'007}
\DeclareMathSymbol{\bsfPhi}{0}{bsfletters}{'010}
\DeclareMathSymbol{\ssfPhi}{0}{ssfletters}{'010}
\DeclareMathSymbol{\bsfPsi}{0}{bsfletters}{'011}
\DeclareMathSymbol{\ssfPsi}{0}{ssfletters}{'011}
\DeclareMathSymbol{\bsfOmega}{0}{bsfletters}{'012}
\DeclareMathSymbol{\ssfOmega}{0}{ssfletters}{'012}

\DeclareMathOperator*{\argmin}{arg\,min}

\newcommand{\qednew}{\nobreak \ifvmode \relax \else
      \ifdim\lastskip<1.5em \hskip-\lastskip
      \hskip1.5em plus0em minus0.5em \fi \nobreak
      \vrule height0.75em width0.5em depth0.25em\fi}

\newcommand{\floor}[1]{\left\lfloor{#1}\right\rfloor}

\DeclareDocumentCommand \ifcond {m m} {%
	{#1} %
	\IfValueT{#2}{\, \middle|\, {#2}}%
}

\DeclareDocumentCommand \P {e{_} g >{\SplitArgument{ 1 }{ @| }}d() g } {%
	\mathbb{P}%
	\IfValueTF{#1}{_{#1}}
		{\IfValueT{#2}{_{#2}}}%
	\IfValueT{#3}{\left(\ifcond#3}%
	\IfValueT{#4}{\, \middle|\, {#4}}%
	\IfValueT{#3}{\right)}%
}

\DeclareDocumentCommand \E {e{_} g >{\SplitArgument{ 1 }{ @| }}o g } {%
	\mathbb{E}%
	\IfValueTF{#1}{_{#1}}
		{\IfValueT{#2}{_{#2}}}%
	\IfValueT{#3}{\left[\ifcond#3}%
	\IfValueT{#4}{\, \middle|\, {#4}}%
	\IfValueT{#3}{\right]}%
}

\definecolor{gray90}{gray}{0.9}

\ifx\nohighlights\undefined

	\newcommand{\msout}[1]{\text{\color{green} \sout{\ensuremath{#1}}}}
	\newcommand{\del}[1]{{\color{green}\ifmmode \msout{#1}\else\sout{#1}\fi}}
\else

	\newcommand{\msout}[1]{#1}
	\newcommand{\del}[1]{#1}
\fi

\newcommand{\hide}[1]{}

\renewcommand{\figurename}{Fig.}
\newcommand{\figref}[1]{\figurename~\ref{#1}}
\graphicspath{{./Figures/}} 
\ifx\diagnoselabel\undefined
	\relax
\else
	\makeatletter
	 \def\@testdef #1#2#3{%
		 \def\reserved@a{#3}\expandafter \ifx \csname #1@#2\endcsname
		\reserved@a  \else
	 \typeout{^^Jlabel #2 changed:^^J%
	 \meaning\reserved@a^^J%
	 \expandafter\meaning\csname #1@#2\endcsname^^J}%
	 \@tempswatrue \fi}
	\makeatother
\fi

\usepackage{graphicx}
\usepackage{tikz}
\usepackage{color}
\usepackage{pgfplots}
\usepackage{cite}
\pgfplotsset{compat=1.5}

\providecommand{\U}[1]{\protect\rule{.1in}{.1in}}

\theoremstyle{definition}

\begin{document}
\title {GFCN: A New Graph Convolutional Network Based on Parallel Flows}
\author{
    Feng~Ji, Jielong~Yang, Qiang~Zhang and Wee~Peng~Tay,~\IEEEmembership{Senior Member,~IEEE}%
 \thanks{This work was supported in part by the Singapore Ministry of Education Academic Research Fund Tier 2 grant MOE2018-T2-2-019 and by A*STAR under its RIE2020 Advanced Manufacturing and Engineering (AME) Industry Alignment Fund – Pre Positioning (IAF-PP) (Grant No. A19D6a0053). The computational work for this article was partially performed on resources of the National Supercomputing Centre, Singapore (https://www.nscc.sg).}%
\thanks{The authors are with the School of Electrical and Electronic Engineering, Nanyang Technological University, 639798, Singapore (e-mail: jifeng@ntu.edu.sg, JYANG022@e.ntu.edu.sg, q.zhang, wptay@ntu.edu.sg). The first three authors contribute equally to this work.}
}

\maketitle

\begin{abstract}
To generalize convolution neural networks (CNN) used for image classification and object recognition, several approaches based on spectral graph signal processing have been proposed. In this paper, we develop a novel approach using parallel flow decomposition of graphs. The essential idea is to decompose a graph into families of non-intersecting one dimensional ($1$D) paths, after which, we may apply a $1$D CNN along each family of paths. We demonstrate that the our method, which we call GFCN (graph flow convolutional network), is able to transfer CNN architectures to general graphs directly, unlike the spectral graph methods. By incorporating skip mechanisms, we show that GFCN recovers some of the spectral graph methods. In addition, GFCN can be used in conjunction with attention mechanisms similar to those in graph attention networks, which have shown better performance than spectral graph methods. To show the effectiveness of our approach, we test our method on the information source identification problem, a news article classification dataset and several vertex-wise classification datasets, achieving better performance than the current state-of-the-art benchmarks based on spectral graph methods and graph attention networks.
\end{abstract}


\section{Introduction}

Suppose $G$ is a graph, which can be weighted and directed. A graph signal on $G$ is a function that assigns a number to each vertex of $G$. The graph describes the relationships amongst the vertex signals, e.g., the relative physical positioning of the vertices or the correlations between the vertex signals. Graph signals An important example of a graph signal an image. In this case, the graph is just a two dimensional ($2$D) lattice, while the graph signals are either RGB values or gray scale values. 

\begin{figure}[!htb]
	\centering
	\includegraphics[scale=0.5]{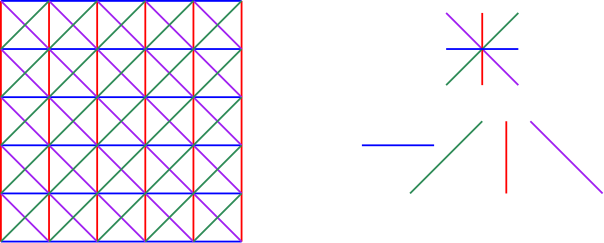}
	\caption{In this example, an image is represented by a $2$D lattice with diagonal connections as shown in (a). The entire graph has four directions, highlighted in different colors. In (b), we show a typical $3\times 3$ filter. It can be decomposed into four $1$D filters, each corresponding to a direction of the lattice in (a).}\label{fig:2Dlattice}
\end{figure}

In recent years, convolution neural networks (CNNs) have been used extensively in a large array of applications (e.g., \cite{Law97,lec98,lec10,Lec15}) with notable success in image processing. There are attempts to extend the CNN architecture to general graphs (e.g., \cite{Hen15,Def16,Kip16,Nie16,Edw16,Du17,Mon17,Ort17,Suc17,Wan18}). Most of these approaches are based on graph signal processing \cite{Shu13, San13, San14, Gad14, Ort18, JiTay:J19} using spectral graph theory and graph Fourier transforms. We would like to recall the main principle of such an approach. For the convolution layer, a polynomial (e.g., a Chebyshev polynomial) of a chosen graph shift operator, such as the graph Laplacian or adjacency matrix, is used as the convolution operator \cite{Def16}. The graph downsampling, as analogous to pooling, usually involves graph clustering. For convenience, we call such an approach a graph convolutional network (GCN). However, the convolution is not analogous to traditional CNN for lattice graphs, and the downsampling step is usually complicated and less canonical. In this paper, we propose a completely different approach that exploits the geometric structure of a graph. To motivate the proposed work, we briefly review some key ingredients of CNN. For convenience, we regard each pixel of an image as a vertex in the graph (shown in \figref{fig:2Dlattice}). Each vertex is connected to the neighboring pixels (including the neighbors on the diagonals). A convolution filter (\figref{fig:2Dlattice}(b)) can be viewed as a signal on a smaller lattice. The convolution operation is performed by taking the Euclidean dot product of the filter placed at various vertices of the image lattice. The filter is used to examine local properties of the image signal, such as existence of edges in different directions. We shall make a comparison of our approach with GCN in Section~\ref{sec:re}. 

An important feature of this approach is that the filter is shared at different places of the image, as the $2$D lattice (for convenience, throughout this paper, when we refer to a 2D lattice, we mean a grid graph with diagonal connections added as shown in \figref{fig:2Dlattice}) is homogeneous and most of the vertices (except those at the image boundary) have the same local structure. However, an attempt to generalize this approach to general graph-structured data faces the immediate difficulty that different vertices may have different number of neighbors and local neighborhoods of each vertex may differ greatly. Such inhomogeneity renders filter sharing difficult to achieve. 

In this paper, we take the point of view that a small filter, such as a $3\times 3$ filter, can be viewed as the superposition or addition of four $1$D filters, each of size $3\times 1$ (cf.\ \figref{fig:2Dlattice}(b)). Each filter operates in a direction of the $2$D lattice. On the other hand, each direction of the $2$D lattice consists of non-crossing paths, or parallel paths. Moreover, each edge of the lattice belongs to one among the four directions. This filtering scheme with shared $1$D filters is readily generalizable to general graphs, as long as we find the correct notion of  ``directions" (called \emph{parallel flows} below) with the properties discussed above. In \figref{gfcn2}, we show some examples, none of which is a lattice. We demonstrate how we can decompose each graph into different ``directions''. Therefore, to design filters, we only need to design a single $1$D filter for each direction. We call our approach the graph flow convolutional network (GFCN).

\begin{figure}[!htb]
	\centering
	\includegraphics[scale=0.47]{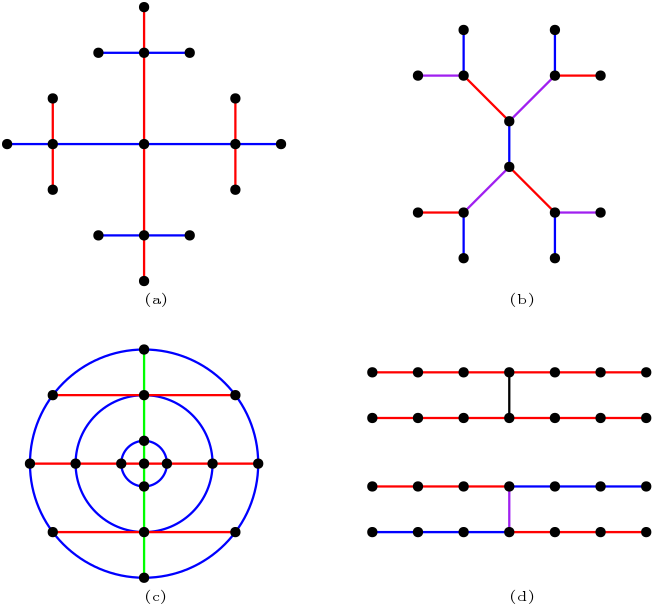}
	\caption{In (a), we have two ``directions'' for the tree, highlighted in blue and red. In (b), we can also cover the tree by two ``directions'', with purple edges common to both directions. In (c), we have a more general graph, and we use three ``directions''. Of course, we may have different ways to give ``directions'' to the same graph, as shown in the simple example in (d).}\label{gfcn2}
\end{figure}

\begin{figure}[!htb]
	\centering
	\includegraphics[scale=0.5]{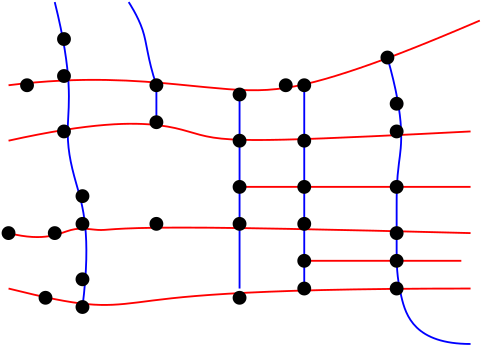}
	\caption{A part of a road network modified from Figure~1 in \cite{Wu16}. The two directions are marked in blue and red.}\label{gfcn11}
\end{figure}

For a concrete example, a traffic network can usually be represented by a graph (an illustration is shown in \figref{gfcn11}) that is non-homogeneous, i.e., there are a lot of different local structures in the graph. However, as suggested in the paper, we can use very few ``directions" to decompose the graph. By restricting to 1D ``directions'', one can systematically perform convolution operations with ease.

Suppose each vertex of the graph is associated with a feature vector. An important complement to graph convolution is the graph attention mechanism (GAT) \cite{Vel18} learns the pairwise interactions between vertices by re-weighting edges in an existing graph and reconstructing vertex feature vectors by using the existing feature vectors. A single GAT layer takes into account the correlations with immediate neighbors of each vertex. By combining both the graph attention mechanism with our proposed GFCN layer(s), we can achieve long spatial correlations over the graph vertex domain and retain the benefits of both graph attention and convolution mechanisms. On the other hand, the GFCN framework allows us to build models by taking analogy from CNN models. An example worth mentioning is the skip layer mechanism in the residual neural network (ResNet) \cite{He16}, which allows us one to build and train deep networks. Furthermore, GFCN can deal with cases where the number of features associated with each vertex is very small or the features are categorical. For example, if each vertex feature is a scalar, GAT is not expected to perform well. We shall discuss how to adopt the attention mechanism as well as the skip layer mechanism for GFCN in Section~\ref{sec:ad}.

The rest of the paper is organized as follows. In Section~\ref{sec:pf}, we introduce the notion of parallel flows and establish the theoretical foundation for the paper. In Section~\ref{sec:pfcnn}, we explain the building blocks of a graph CNN model using parallel flows and discuss how to build a GFCN model. In addition, we also indicate how to include more sophisticated mechanisms, and relate GFCN with the popular GCN model. We present experimental results on different applications in Section~\ref{sec:exp} and conclude in Section~\ref{sec:con}.

\section{Parallel flows} \label{sec:pf}

In this section, we introduce the concept of parallel flows and show that a graph can be decomposed into 1D parallel flows, the number of which is bounded by a function only of the maximal degree of the graph. Therefore, the number of flows does not scale with the size of the graph.

\subsection{1D parallel flows}

We assume that $G=(V,E)$ is a connected, undirected, unweighted simple graph. We shall describe how to remove such restrictions in building up a graph CNN model in the next section. 

We assume that a subset of vertices $V'\subset V$ is fixed. An intuitive interpretation is that $V'$ serves as a choice of the \emph{boundary} of $V$. Moreover, $V'$ can be the empty set. Examples of $V'$ are as follows.
\begin{Example}\
	\begin{enumerate}[(a)]
		\item A typical example is when $G$ is a tree and $V'$ consists of the degree $1$ vertices or leaves of $G$. 
		\item If $G$ is a finite 2D lattice like an image, then there is a natural choice of the boundary $V'$ as the vertices with degree less than $8$. 
	\end{enumerate}
\end{Example}

\begin{Definition}\
\begin{enumerate}[(a)]
	\item A \emph{path} $P$ of $G$ is a connected subgraph of $G$ such that each vertex of $P$ has degree not more than $2$ in $P$. In particular, a single vertex or a closed loop is a path.
	\item A path $P$ is \emph{non-extendable} if either $P$ is a single vertex, a closed loop or the ends of $P$ belong to the set $V'$. We allow the interior of $P$ to contain vertices of $V'$.
	\item Two paths $P_1$ and $P_2$ are said to be \emph{parallel} to each other if they do not have common vertices and edges in $G$. 
\end{enumerate}	
\end{Definition}

The following key concepts allow us to generalize the notion of ``directions" in $2$D lattices (as discussed in the introduction). 

\begin{Definition}
A set of paths $\mathcal{P} = \{P_1,\ldots, P_n\}$ is called a \emph{parallel flow} if:
\begin{enumerate}[(a)]
\item $P_1, \ldots, P_n$ are non-extendable, and
\item $P_1, \ldots, P_n$ are pairwise parallel.
\end{enumerate}
\end{Definition}

\begin{Definition}
	Given an $\epsilon\in (0,1]$, a set of parallel flows $\mathcal{P}_1,\ldots, \mathcal{P}_m$ is \emph{an $\epsilon$-cover} of $G$ if the union of all the paths in all the parallel flows contains $\epsilon$ fraction of all the edges of $G$. If $\epsilon = 1$, we simply call any $1$-cover a cover. The smallest $m$ such that there is an $\epsilon$-cover consisting of $m$ parallel flows is denoted as $\mu(G,V',\epsilon)$. We abbreviate $\mu(G,V',1)$ as $\mu(G,V')$. 
\end{Definition}

It is sometimes of practical use to have $\epsilon<1$. For example, in \figref{gfcn2}(d), if we allow $\epsilon =0.9$, then we only need to use one parallel flow (the two red horizontal paths) to obtain a $0.9$-cover. 

We next show some general properties regarding $\epsilon$-covers. Let $d(\cdot,\cdot)$ be the metric of the graph $G$ and $(u,v)$ be the shortest path in $G$ between vertices $u$ and $v$, excluding the end vertices.

\begin{Lemma}\ \label{lem:ivsv}
	\begin{enumerate}[(a)]
		\item\label{it:sgss} If $V' \subset V''$, then $\mu(G,V'',\epsilon)\leq \mu(G,V',\epsilon)$.
		\item \label{it:sgia} Suppose $G_1=(V_1,E_1)$ is a subtree of a tree $G_2=(V_2,E_2)$ and $V_i'\subset V_i$, for $i=1,2$. Moreover, for each $v \in V_2'$, the vertex in $V_1$ closest to $v$ belongs to $V_1'$. Then $\mu(G_1,V_1')\leq \mu(G_2,V_2')$.
		\item \label{it:sgga} Suppose $G=G_1\cup G_2$ and $V_1'\cup V_2'\subset V'$. Then $\mu(G,V')\leq \mu(G_1,V_1')+\mu(G_2,V_2')$.
	\end{enumerate}
\end{Lemma}
\begin{IEEEproof}\
\begin{enumerate}[(a)]
	\item Since $V'\subset V''$, any non-extendable path \gls{wrt} $V'$ is also not extendable \gls{wrt} $V''$. Therefore any $\epsilon$-cover of $G$ \gls{wrt} $V'$ is also an $\epsilon$-cover \gls{wrt} $V''$, hence the claim follows.
	\item Let $P$ be a non-extendable path of $G_2$. Since $G_2$ is a tree, $P$ is not a closed loop. Consider $P'=P\cap G_1$, a sub-path of $P$. We claim that $P'$ is a non-extendable path of $G_1$. If $P'$ is empty, the claim follows trivially. Suppose $P'$ is non-empty. Let $u\in P'$ be an end vertex of $P'$. If $u\in V_2'$, then since $u\in V_1$, the vertex $u\in V_1'$ by our assumption. On the other hand, suppose $u\notin V_2'$. Let $v\in P$ be the end vertex of $P$ such that the path $(u,v) \notin P'$. If there exists a $u'\in V_1$ with $u'\ne u$ such that $d(u',v) < d(u,v)$, then since $G_1$ is a tree, there exists a path in $G_1$ connecting $u'$ and $u$ and hence $u$ cannot be an end vertex. This shows that $u$ is closest to $v$ in $V_1$ and by our assumption, it belongs to $V_1'$. Therefore, $P'$ is non-extendable. It is clear that being parallel is preserved by taking intersection with $G_1$. Therefore, by taking intersection, a cover of $G_2$ by parallel flows yields a cover of $G_1$. The inequality now follows.
	\item Given two collections of parallel flows $\{\mathcal{P}_1, \ldots, \mathcal{P}_m\}$ and $\{\mathcal{Q}_1, \ldots, \mathcal{Q}_n\}$, define their union to be the collection of parallel flows $\{\mathcal{P}_1, \ldots, \mathcal{P}_m, \mathcal{Q}_1, \ldots, \mathcal{Q}_n\}$. Any union of covers  of $G_i$, $i=1,2$, by parallel flows is also a cover of $G$ by parallel flows, under the assumption that $V_1'\cup V_2'\subset V'$. Hence $\mu(G,V')\leq \mu(G_1,V_1')+\mu(G_2,V_2')$.
\end{enumerate}
\end{IEEEproof}

\begin{Theorem} \label{thm:sgia}
	Suppose $G=(V,E)$ is a tree. Let $d_{\max}$ be the maximal degree of $G$ and $V'$ consist of vertices with degree strictly smaller than $d_{\max}$. Then $\mu(G,V')=\floor{(d_{\max}+1)/2}$.
\end{Theorem}

\begin{IEEEproof}
See Appendix~\ref{sec:pt}.
\end{IEEEproof}

As an immediate consequence, we have that if $G=(V,E)$ is a tree, then $\mu(G,V)= \floor{(d_{\max}+1)/2}$ by \cref{lem:ivsv}\ref{it:sgss} as $V$ clearly contains all vertices with degree degree strictly smaller than $d_{\max}$.

Now for a general graph $G$, we can always find a spanning subtree $G_1$ of $G$ (for example a breadth-first-search spanning tree from any non-isolated vertex). The maximal degree of $G_1$ is clearly no greater than $d_{\max}$. Moreover, if we let $G_2$ be the (possibly disconnected) subgraph of $G$ by first removing those edges contained in $G_1$ and then isolated vertices, the maximal degree of $G_2$ is strictly smaller than $d_{\max}$. Therefore, a simple induction yields the following estimation for a general graph.

\begin{Corollary} \label{coro:fagg}
	For any graph $G=(V,E)$ (which can be disconnected), let $d_{\max}$ be the maximal degree of $G$. Then $\mu(G,V) \leq (\floor{(d_{\max}+1)/2}+1)\floor{(d_{\max}+1)/2}$.
\end{Corollary}

\subsection{Product graph and high dimensional generalization}

Although in this paper we mainly focus on $1$D parallel flows, as they are omnipresent in any graph, we still briefly discuss its high dimensional generalization. Recall that given two graphs $G_1=(V_1,E_1)$ and $G_2 = (V_2, E_2)$, their \emph{product} $G = G_1\times G_2$ is defined as follows: the vertices of $G$ consists of pairs $(v_1,v_2)$ with $v_1\in V_1, v_2 \in V_2$. Two pairs $(v_1,v_2)$ and $(v_1',v_2')$ are connected by an edge if $v_1=v_1', (v_2, v_2') \in E_2$ or $v_2=v_2', (v_1,v_1') \in E_1$. 

If $P_i$ is a path on $G_i$ for $i=1,2$, then $P_1\times P_2$ is a $2$D lattice in $G$. Therefore, on $G$, we may construct collections of non-intersecting $2$D lattices as 2D analogy to $1$D parallel flows, or $2$D ``parallel flows". Notions such as \emph{cover} and \emph{$\epsilon$-cover} have obvious 2D counterparts. A simple yet useful example is any $2$D lattice can be taken as the product of two paths. It has an obvious cover by a single $2$D ``parallel flow", which is the entire graph itself.   

The construction can be repeated if a graph $G$ is an iterated product of subgraphs, yielding higher dimensional counterparts.

\begin{figure}[!htb]
	\centering
	\includegraphics[scale=0.55]{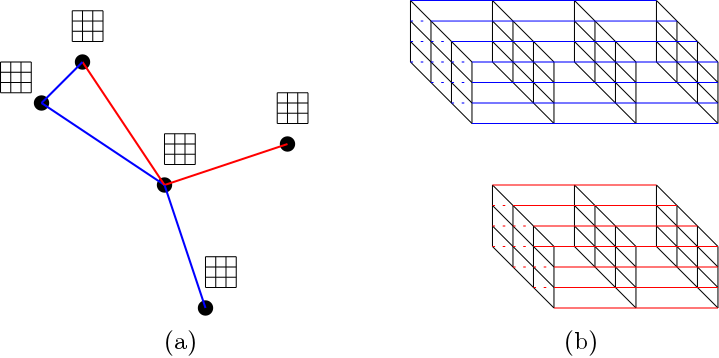}
    \caption{Illustration of Example~\ref{eg:ane}.}\label{gfcn10}
\end{figure}

\begin{Example} \label{eg:ane}
As an example (illustrated by \figref{gfcn10}), suppose the graph $G$ is a surveillance network consisting of $5$ cameras as in (a). A parallel flow decomposition results in two flows containing a single path each, colored in blue ($P_1$) and red ($P_2$) respectively. If each camera takes a picture, which is represented by a grid $H$, the aggregated data can be viewed as signals on the product graph $G\times H$. As discussed in this subsection, we may analyze the data using two $3$D flows consisting of $P_1\times H$ and $P_2\times H$ respectively (as shown in \figref{gfcn10}(b)). To perform convolution, one needs to apply $3$D convolution filters, with $2$ coordinates for the image and $1$ coordinate for the graph component. 
\end{Example}

\section{GFCN: a parallel flow based graph convolutional network} \label{sec:pfcnn}

In this section, we propose a convolutional network framework for graph-structured data based on the concept of parallel flow decomposition of a graph introduced in the previous section. For a graph, recall that \emph{a graph signal or a graph label} is a function that assigns a (real) number to each vertex $v\in V$ \cite{Shu13, San13, San14, Gad14, Ort18, JiTay:J19}. On each vertex $v$ of a path in a parallel flow, it retains the signal of the corresponding vertex in the original graph.

\subsection{Architectural components}

For a fixed $0<\epsilon \leq 1$, let $\{\mathcal{P}_1,\ldots,\mathcal{P}_m\}$ be an $\epsilon$-cover of the graph $G=(V,E)$ by parallel flows w.r.t.\ $V'=V$. To build up a GFCN (exemplified in \figref{gfcn12}), we have the following components:

\begin{figure}[!htb]
	\centering
	\includegraphics[width=0.8\textwidth]{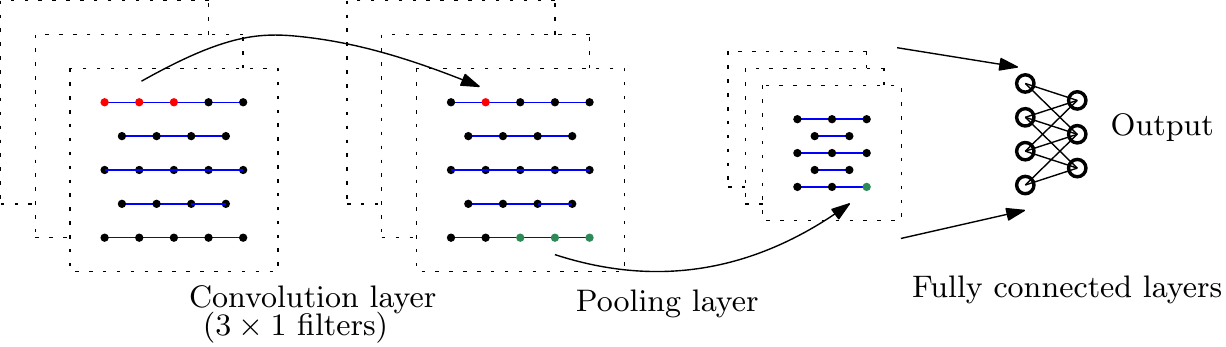}
	\caption{An illustration of a GFCN model with a convolution layer, a pooling layer and fully connected layers.}\label{gfcn12}
\end{figure}    

\begin{enumerate}[(a)] 
\item {\bf Convolution layers:} As each parallel flow $\mathcal{P}_i$ is the union of linear paths, we may apply a \emph{common} set of standard 1D filters for each path in the flow. We briefly recall that a filter is specified by the parameters: $p$ as the padding size, $n$ as the filter size, $s$ as the stride number and $c$ as the number of channels. It is preferable that $n$ is an odd number such that there is a unique center.

\item {\bf Pooling layers:} Similar to the convolution layers, the pooling layer is the standard 1D pooling specified: by $n$ (preferably being odd) the pooling size, and $s$ the stride number. We usually use max pooling, though average pooling is also a viable option. 

\item {\bf Fusion layers:} This is a new type of architectural component to establish communication among different parallel flows. We fix an ordering of the vertices of $G$, and each vertex has a unique index. For each vertex of the linear paths in the parallel flows, we record the index of the vertex in the graph $G$. A vertex in $G$ might appear in different parallel flows, and all of them have the same index. Each convolution or pooling operation centered at vertex $v$ makes $v$ retain its index. In the fusion layer, we apply a fusion function $f$ across all the vertices with the same index in all the parallel flows (see \figref{gfcn6} for an example). Preferred choices of $f$ include the $\max$ function, the average function and the sum function. 
\begin{figure}[!htb]
	\centering
	\includegraphics[scale=0.5]{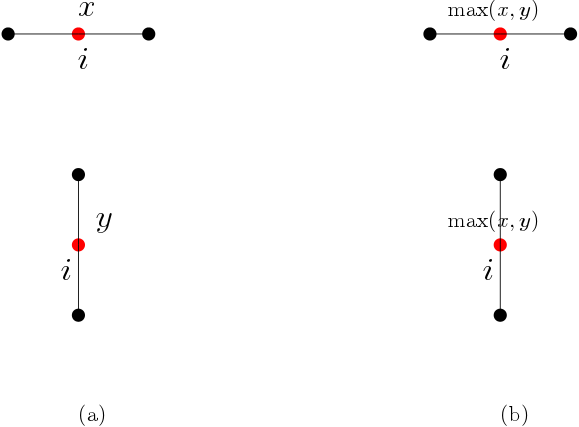}
	\caption{In (a), both the central vertex of the horizontal and vertical paths are indexed by $i$ with signals $x$ and $y$ respectively. After the fusion layer with $\max$ fusion function $f$, they both have the common signal $\max(x,y)$.}\label{gfcn6}
\end{figure}    

We remark that the the fusion layer is designed so that we are able to formally recover a $2$D pooling layer as a concatenation of a $1$D pooling layer and a fusion layer. However, experiments suggest that the usage of $1$D pooling layers alone might be sufficient, and the additional involvement of fusion layers does not add much to the performance.    

\item {\bf Fully connected layers:} These layers are again standard. They are the last few layers of a architectural pipeline when the dimension of the feature vector is sufficiently small.
\end{enumerate}

For certain graphs, $\epsilon$-covers can be obtained by inspection. For example, for the $2$D lattice, we may use the canonical cover $\{\mathcal{P}_1,\mathcal{P}_2\}$, where $\mathcal{P}_1$ and $\mathcal{P}_2$ consist of horizontal and vertical paths respectively. If $G$ is obtained from the $2$D lattice by connecting vertices that are diagonally adjacent, we may further include parallel flows $\mathcal{P}_3$ and $\mathcal{P}_4$ consisting of the paths in the two diagonal directions respectively.

In the cases where a parallel flow decomposition cannot be obtained by inspection, the discussions in Section~\ref{sec:pf} (for example, the proof of Corollary~\ref{coro:fagg}) tell us how we may proceed. More precisely, let $\mathcal{S}$ be an operation that take $G$ and $v\in V$ as inputs and a spanning tree $T$ of $G$ as an output. Examples of $\mathcal{S}$ include the breadth-first-search (BFS) and the depth-first-search (DFS) algorithms. W may choose a vertex $v \in V$ and generate a spanning tree $T = \mathcal{S}(G,v)$. By removing the edges belonging to $T$ from $G$, we obtain a (possibly disconnected) subgraph $G'$ of $G$ with \emph{strictly smaller} number of edges. On the other hand, the steps in the proof of Theorem~\ref{thm:sgia} can be used to decompose a tree into parallel flows. To be more precise, we may label the edges using numbers in the set $\{1,\ldots, \floor{(d_{\max}+1)/2}\}$ such that for each node $v$, every label is used at most twice for edges connected to $v$. Subsequently, the edges with the same label are collected to form one parallel flow. This procedure can be repeated until a sufficient number of edges are included in the parallel flows.

Once an $\epsilon$-cover of $G$ by parallel flows is obtained, there is an option to regularize the size of the parallel flows by removing short paths and sub-dividing long paths into shorter ones.

The rest of the steps can be derived by modifying any CNN model. For the applications of this paper, we mainly use variations of the following simple model: input layer $\to$ convolution layer $\to$ pooling layer $\to$ fusion layer $\to$ convolution layer $\to$ pooling layer $\to$ fusion layer $\to$ fully connected layers $\to$ output layer. 

If the operation $\mathcal{S}$ in the parallel flow decomposition can be applied to directed graphs, so can GFCN be adapted to such graphs. Moreover, GFCN can be applied to weighted graphs as well, as we only need to form the weighted dot product with edge weights in the convolution layers.

\subsection{Comparison between GFCN and CNN}

We make a comparison between GFCN and CNN. To emphasize the similarities between them, we compare them side-by-side in the following table, which also serves as a dictionary between these two frameworks.

\begin{table}[!htb]
	\caption{Comparison between CNN and GFCN.} \label{tab:1}
	\centering  
	\scalebox{1.2}{
	\begin{tabular}{|l|c|c|}  
		\hline
		\emph{Components} &  \emph{CNN} & \emph{GFCN}\\ 
		\hline \hline 
		Convolution & $2$D filters & Multiple $1$D filters\\
		\hline
		Down sampling & $2$D pooling & $1$D pooling $\&$ fusion \\ 
		\hline
		Striding & $2$D striding & $1$D striding \\
		\hline
	\end{tabular}}
\end{table}
  
From Table~\ref{tab:1}, we see that GFCN bears much resemblance to CNN, while GFCN has the advantage of being generalizable to arbitrary graphs. Moreover, we expect that more sophisticated CNN models can be ``translated" into GFCN models using Table~\ref{tab:1}.

We perform a simple demonstration using the MNIST dataset.\footnote{http://yann.lecun.com/exdb/mnist/} The parallel flow decomposition can be obtained by inspection, namely, the parallel flows consist of horizontal paths, vertical paths and paths in the two diagonal directions. We build a model with two convolution layers each with $16$ channels of size $3$ $1$D filters, with stride $1$. Immediately after each convolution layer, there is a $1$D pooling layer of size $3$, with stride $2$. The last two layers are fully connected layers. In summary, we have: input layer $\to$ convolution layer $\to$ pooling layer $\to$ convolution layer $\to$ pooling layer $\to$ $2$ fully connected layers $\to$ output layer. This very simple model is able to achieve a \emph{$98.7\%$ accuracy}, which is comparable with accuracy achievable by CNN architectures such as \emph{$98.3\%$ by LeNet-1} and \emph{$98.9\%$ by LeNet-4} \cite{lec10}.

\subsection{Advanced mechanisms} \label{sec:ad}

In this subsection, we describe how the skip layer mechanism and attention mechanism can be added to GFCN. 

\begin{figure}[!htb]
	\centering
	\includegraphics[scale=1]{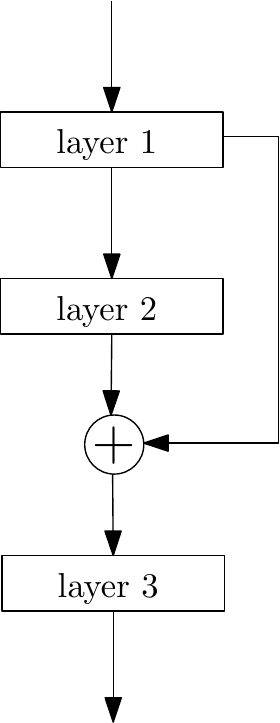}
	\caption{An illustration of the skip layer mechanism: there is a direct connection from layer $1$ to layer $3$ that skips layer $2$. The output of layer $2$ and layer $1$ are summed up to give the input of layer $3$.}\label{gfcn7}
\end{figure}

By using Table~\ref{tab:1}, we may translate a CNN architecture, with minor modifications, to obtain a GFCN architecture. First, we describe the skip layer mechanism, which is the key component of residual network (ResNet) \cite{He16,Sri15}. The skip layer mechanism allows one to train a deep network as in the traditional CNN. Similar to the ResNet, the skip layer mechanism allows a direct connection from a layer to another layer that is not the immediate next (see \figref{gfcn7} for an illustration). We may take a direct summation between source and target layers along the connection that skips layers. Backward propagation learning follows the exact same formula as ResNet. Moreover, there can be multiple skip layer connections emanating from the same source or terminating at the same target.

\begin{figure}[!htb]
	\centering
	\includegraphics[scale=0.45]{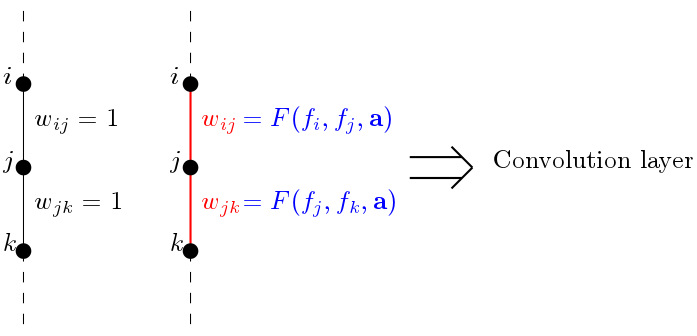}
	\caption{An illustration of the attention mechanism.}\label{gfcn9}
\end{figure}

In certain problems with unweighted graph, each edge captures correlation between two vertices. To give a quantitative reconstruction, one may need to learn an edge weight for each edge of the graph. This can be achieved using the attention mechanism. We now explain how it can be welded into the GFCN framework (see \figref{gfcn9} for an illustration).   

We assume that for vertex $i$, there is an associated feature vector $f_i$ (of dimension independent of $i$). In the attention layer, there is a function $F(f_i,f_j,{\bf a})$ parametrized by a vector ${\bf a}$, taking a pair of features $f_i$ and $f_j$ as inputs. The vertices $i$ and $j$ are \emph{a priori} connected by an edge. In a parallel flow, the parameters ${\bf a}$ are shared and hence independent of vertices $i$ and $j$. A particularly important example of $F$ is the output of a neural network, while ${\bf a}$ are the weights and biases of the neural network. In the subsequent GFCN convolution layer, $w_{ij} = F(f_i,f_j,{\bf a})$ are used as new edge weights. The parameters ${\bf a}$ are usually trained together with parameters such as convolution coefficients and parameters in the fully connected layers through backpropagation. Effectively, $F$ re-evaluates edge weights of the paths in flows and hence does not change the size of flows.

\subsection{GCN revisited} \label{sec:re}

One of the most popular graph CNN architectures is based on spectral graph signal processing (e.g., \cite{Shu13,Hen15,Def16,Kip16,Edw16,Du17,Mon17,Ort17,Suc17,Wan18}), abbreviated by GCN for convenience. Both GCN and GFCN can be useful in different applications. We briefly recall that a GCN convolution filter is usually a polynomial of a chosen graph shift operator. For a single convolution layer of GFCN, a filter assigns different weights to the neighbors of the central vertex in the convolution operation. On the other hand, if the graph adjacency or Laplacian matrix is used as the graph shift operator for GCN, then a filter assigns a \emph{common} weight to all the neighbors of the central vertex and vertices receive different weights only when their distances to the central vertex are different. In this respect, GFCN is a closer relative to CNN.     

In addition, if multiple layers and the skip mechanism are used, one may use a GFCN model to produce any convolution layer in GCN. To formalize this statement, we use $S$ to denote a fixed graph shift operator, which is the graph adjacency matrix, the graph Laplacian matrix or their normalized versions. Let $p$ be a polynomial and hence $p(S)$ is graph convolution filter, i.e., if $X$ is a graph signal, the convolution layer with $p$ gives $p(S)X$ as the output (before any nonlinearity).

\begin{Proposition} \label{prop:sai}
	Suppose no edge is contained in different flows inside a given set of parallel flows. If $n$ is the degree of the polynomial $p$, then there is a GFCN model with $2(n-1)$ hidden layers producing the exact same output as the GCN convolution filter $p(S)$ for any input graph signal.
\end{Proposition}
\begin{IEEEproof}
See Appendix~\ref{sec:pp}.
\end{IEEEproof}

\section{Numerical Experiments} \label{sec:exp}

In this section, we apply the GFCN model to a few learning problems involving graph-structured data. We present results on a synthetic information source identification dataset, the 20news dataset \cite{Joa97}, three widely used citation network datasets Citeseer, Cora and Pubmed \cite{Sen08, Kip16, Vel18}, and the protein-protein interaction (PPI) dataset \cite{Les17}. We compare the performance of GFCN with various state-of-the-art classifiers proposed in the literature for each of these datasets.

\subsection{Information source identification}

In this section, we study the information source identification problem with snapshot observations \cite{Shah2010,Luo2013,LuoTay2013,Zhu2016,Ji17} using synthetic data. A piece of information or rumor is spread over a network, represented as a graph, from a source vertex. Any vertex that acquires the information is called infected. It has a positive probability $p_i$ to infect any of its neighbors. Moreover, each infected vertex has a positive probability $p_r$ to recover from the infected status. The model is called the SIRI model.

A snapshot observation of the infection status of all the vertices are made (usually when not all the vertices are infected). For each vertex, only whether the vertex is infected or un-infected is known. Our goal is to identify the source from the snapshot observation. 

For the simulation, we consider the Enron email network with $500$ vertices.\footnote{https://snap.stanford.edu/data/email-Enron.html} For each sample, we randomly choose a source and generate a spreading with random, unknown $p_i$ and $p_r$. A snapshot observation is made with about $20\%$ of the total population being infected. We use $8000$ samples for training and $1000$ samples for testing.

We perform BFS based flow decomposition and run GFCN with two convolutional layers. The output of GFCN gives each vertex a score indicating its likelihood of being the source. The performance is evaluated using $x\%$ accuracy, i.e., the source belongs to the set of vertices with top $x\%$ likelihood scores. We compare the performance of GFCN with the Jordan center method \cite{Luo2017}, which is effectively a feature engineering approach. Briefly, the method finds a Jordan center $s$ of the infected vertices $I$ as the source, i.e.,
\begin{align*}
s = \argmin_{v \in G} \max_{u\in I} d(u,v), 
\end{align*} 
where $d(\cdot,\cdot)$ is the distance function on the graph $G$ and $I$ is the set of infected vertices. In this example, each vertex signal is a categorical variable indicating whether the vertex is infected or not. The GAT approach cannot be readily applied in this situation.

The results are summarized in Table~\ref{tab:4}. From the results, we see the potential of GFCN as compared with the Jordan center method. However, the Jordan center method is unsupervised and does not require any training samples.

\begin{table}[!htb]
	\caption{Performance of GFCN and Jordan center.} \label{tab:4}
	\centering  
	\scalebox{1.3}{
		\begin{tabular}{|l|c|c|c|c|}  
			\hline
			\emph{Method} & \emph{$1\%$-accuracy} & \emph{$5\%$-accuracy} & \emph{$10\%$-accuracy}\\ 
			\hline\hline
			Jordan center & $9.79\%$ & $20.73\%$ & $33.26\%$  \\
			\hline
			GFCN & $\bm{44.87\%}$ & $\bm{74.26\%}$ & $\bm{85.42\%}$  \\
			\hline
	\end{tabular}}
	
\end{table}

\subsection{News article classification}

We next apply GFCN to news article classification on the publicly available 20news dataset \cite{Joa97}. The graph signals are constructed as in \cite{Def16}: each document $x$ is represented by a normalized bag-of-words model and the underlying graph $G$ (of size $1000$) is constructed using a 16-NN (nearest neighbor) graph on the word2vec embedding \cite{Mik13} with the $1000$ most common words (instead of $10,000$ keywords as in \cite{Def16}). There are $16,617$ texts with each label from one amongst $20$ categories. 

The decomposition of $G$ into parallel flows makes use of both BFS and DFS based heuristics. We remove all paths with length smaller than $6$. There are $43$ parallel flows that remain, and they form a $0.91$-cover of $G$. In the model, for the pooling layer, size $2$ max pooling is used with stride $1$.

Along with the test accuracy of GFCN, we also include the performance of a few classification methods from \cite{Def16}, including:
\begin{itemize}
	\item GC32:  GCN model with $32$ feature maps for a convolution layer, and the convolution filter is a polynomials of the normalized graph Laplacian;
	\item FC2500: a neural network with a single hidden layer with $2500$ hidden units;
	\item FC2500-FC500: the neural network with two hidden layers with $2500$ and $500$ hidden units respectively.
	\item Softmax: the neural network with a single softmax layer.
\end{itemize}
Readers are referred to \cite{Def16} for the detailed model architectures. The results are show in Table~\ref{tab:2} middle column. We see that GFCN outperforms all the other methods.

\begin{table}[!htb]
	\caption{Performance of GFCN and other models.} \label{tab:2}
	\centering  
	\scalebox{1.2}{
		\begin{tabular}{|l|C{1.5cm}|C{2cm}|}  
			\hline
			\emph{Method} & \emph{Accuracy (full)} &  \emph{Accuracy (non-sparse)} \\ 
			\hline \hline 
			GFCN & \textbf{64.5\%} & \textbf{70.0\%}\\
			\hline
			GC32  & 62.8\% & 67.1\% \\
			\hline
			FC2500  & 63.2\% & 66.5\% \\
			\hline
			FC2500-FC500  & 63.3\% & 67.7\% \\ 
			\hline
			Softmax & 61.0\% & 67.3\% \\
			\hline
	\end{tabular}}
\end{table}   

For the results, we notice that many errors occur for those texts with sparse graph signals, i.e., most of the components are $0$. On the other hand, some of the class labels are similar and hard to distinguish for texts with sparse graph signals, e.g., ``soc.religion.Christian" and ``talk.religion.misc". Therefore, we perform another test on the sub-dataset with $12000$ texts by removing those texts whose graph signals has less than $13$ nonzero components. The results are summarized in Table~\ref{tab:2} (right column). Again, GFCN performs the best.

\subsection{Vertex-wise labeling }

In the previous experiment on news article classification, each document is represented by a single graph assigned with a single overall label. In this subsection, we consider the problem of vertex-wise labeling. We use the citation network datasets: Citeseer, Cora  and Pubmed \cite{Sen08, Kip16, Vel18}. We briefly recall that each dataset contains bag-of-words representation of documents, which are used as feature vectors. The documents are connected by citations links, forming the citation networks. The documents are divided into a few categories, while a small percentage of them are labeled. The task is to label all the documents. The details of the datasets can be found in \cite{Sen08}. This is known as transductive learning as the training and testing graphs are the same. 

We also test on the PPI dataset \cite{Les17} that consists of graphs corresponding to different human tissues. The dataset is inductive, which means that training and testing graphs are different and testing graphs are unobserved during training. 

As the problems considered in this subsection are semi-supervised and the graph of each dataset is disconnected, we need to modify the flow decomposition scheme. Fix a number $l$ (preferably $5$ or $7$). The flows are formed by taking paths of length $l$ centered at vertices with labels. For each layer, a single set of filters are used with multiple channels. In addition, to make use of the feature vector of each vertex effectively, we deploy the attention mechanism. More precisely, after the input layer, we insert two attention layers before each of two $1$D convolution layers, followed by fully connected layers.

We compare the performance of GFCN with effective state-of-the-art methods, including:
\begin{itemize}
	\item Planetoid\cite{Yan16}: semi-supervised classification algorithm based on graph embedding;
	\item ICA\cite{Lu03}: an iterative classification algorithm using link features and statistics;
	\item GCN\cite{Kip16}: multi-layer graph convolutional network with polynomial filters of the normalized graph Laplacian;
	\item GAT\cite{Vel18}: graph attention network with two attention layers and each attention layer evaluates edge weights with pairs of vertex features;
	\item (GAT-)GRevNet\cite{Liu19}: a reversible graph neural network model and GAT-GRevNet is GRevNet implemented with attention mechanism.\footnote{As reported in \cite{Liu19}, the results for Cora and Pubmed are using GAT-GRevNet, and the result for PPI is using GRevNet.}
\end{itemize}
Readers are referred to the above mentioned papers for the detailed model architectures. The results (accuracy for Citeseer, Cora and Pubmed, and F1 score for PPI) are summarized in Table~\ref{tab:3}.\footnote{Some fields in Table~\ref{tab:3} are missing either because the result is not reported in the literature or the method is not applicable to the dataset.} We observe that GFCN has the best performance. 

\begin{table}[!htb]
	\caption{Performance of GFCN and other models.} \label{tab:3}
	\centering  
	\scalebox{1.3}{
		\begin{tabular}{|l|C{1.5cm}|C{1.5cm}|C{1.5cm}|C{1cm}|}  
			\hline
			\emph{Method} & \emph{Citeseer (Accuracy)} & \emph{Cora (Accuracy)} & \emph{Pubmed (Accuracy)} & \emph{PPI (F1)}\\ 
			\hline  \hline 
			GFCN & $\bm{73.7\%}$ & $\bm{83.7\%}$ & $\bm{79.1\%}$ & $\bm{97.5\%}$\\
			\hline
			Planetoid & $64.7\%$ & $75.7\%$ & $ 77.2\%$ & --\\
			\hline
			ICA  & $69.1\%$ & $75.1\%$ & $ 73.9\%$ & --\\ 
			\hline
			GCN & $70.3\%$ & $81.5\%$ & $ 79.0\%$ & $78.0\%$\\
			\hline
			GAT & $72.5\%$ & $83.0\%$ & $ 79.0\%$ & $96.0\%$\\
			\hline
			(GAT-)GRevNet & -- & $82.7\%$ & $ 78.6\%$ & $76.0\%$\\
			\hline
	\end{tabular}}
\end{table}    

\section{Conclusion} \label{sec:con}

In this paper, we introduced a new convolution neural network framework called GFCN for general graph-structured data, based on the idea of decomposing a graph into parallel flows. This approach allows us to mimic CNN architectures already developed for $2$D lattices. We presented a few applications to demonstrate the effectiveness of the approach. For future work, we shall explore more systematic ways for flow decomposition and design deeper GFCN models.

\appendices

\section{Proof of Theorem~\ref{thm:sgia}} \label{sec:pt}

The inequality $\mu(G,V')\geq \floor{(d_{\max}+1)/2}$ is clear. This is because at any vertex $v$ with maximal degree, at least $\floor{(d_{\max}+1)/2}$ paths are required to cover all the neighboring edges. Moreover, none of them are parallel, and hence $\mu(G,V')\geq\floor{(d_{\max}+1)/2}$. We next prove $\mu(G,V')\leq \floor{(d_{\max}+1)/2}$ by starting with some special cases. 

\begin{Lemma_A}\label{lem:regular_tree}
Suppose that the tree $G$ is regular, i.e., every vertex except the leaves have the same degree $d_{\max}$. If $V'$ is the set of leaves of $G$, then $\mu(G,V')=\floor{(d_{\max}+1)/2}$.
\end{Lemma_A}

\begin{IEEEproof}
We proceed by considering the cases where $d_{\max}$ is either even or odd.

\begin{enumerate}[\parindent, itemindent=3\parindent, label={\bf {Case} \arabic*:}, ref={Case \arabic*}]
\item $d_{\max}$ is even.
The case $d_{\max} = 2$ is trivial. We assume $d_{\max}\geq 4$. Starting from a fixed vertex $v_0$ with degree $d_{\max}$, we color its adjacent edges in pairs using distinct colors from $\{1,\ldots, d_{\max}/2\}$. For each neighbor $v$ of $v_0$, we arbitrarily choose an edge adjacent to $v$ that is not the edge $(v_0,v)$ and color it the same color as $(v_0,v)$. We can again color the remaining adjacent edges of $v$ in pairs using distinct colors different from $(v_0,v)$. This procedure can be continued for all the vertices without any conflict as $G$ is a tree. For each color $k$, the union $\mathcal{P}_k$ of edges with the same color forms a parallel flow (see \figref{gfcn3} for an example) because of the following:
\begin{enumerate}[\parindent, label=(\alph*)]
\item For each vertex, there are at most two adjacent edges with the same color. Hence any two distinct paths of $\mathcal{P}_k$ do not intersect.
\item Moreover, if $v$ has degree $d_{\max}$, then it is of degree $2$ in a path in $\mathcal{P}_k$. Hence each path of $\mathcal{P}_{k}$ is non-extendable \gls{wrt} $V'$.
\end{enumerate}
The union of $\mathcal{P}_k,1\leq k\leq d_{\max}/2$ clearly covers $G$. This proves that $\mu(G,V')\leq d_{\max}/2$ as we have presented one such cover.

\begin{figure}[!htb]
\centering
\includegraphics[scale=0.6]{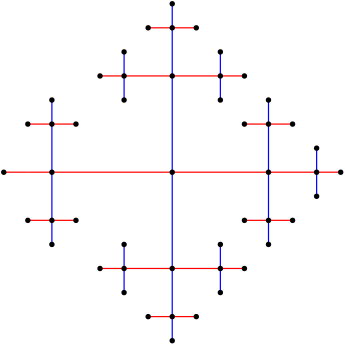}
\caption{In this example, we have $d_{\max}=4$. We label the tree using two colors and obtain a cover with two parallel flows.}\label{gfcn3}
\end{figure}

\item\label{casedmax_3} $d_{\max}=3$. 
We want to show that $\mu(G,V')=2$. We prove this by induction on $|V|$. The base case $|V|=4$ is trivial. For $|V|>4$, we are able to find a subtree $G_1=(V_1,E_1)$ (see \figref{gfcn4}) of $G$ such that  the following holds:
\begin{enumerate}[\parindent, label=(\alph*)]
\item $G_1$ also belongs to Case 2 with $|V_1|=|V|-2$ and $|V_1'|=|V|-1$.
\item There is a vertex $v$ of $G_1$ of degree $1$ such that $G$ is obtained from $G_1$ by attaching two edges $e_1=(v_1,v)$ and $e_2=(v_2,v)$. Furthermore, $V_1' = V'\backslash\{v_1,v_2\}\cup\{v\}$.
\end{enumerate}

\begin{figure}[!htb]
\centering
\includegraphics[scale=0.35]{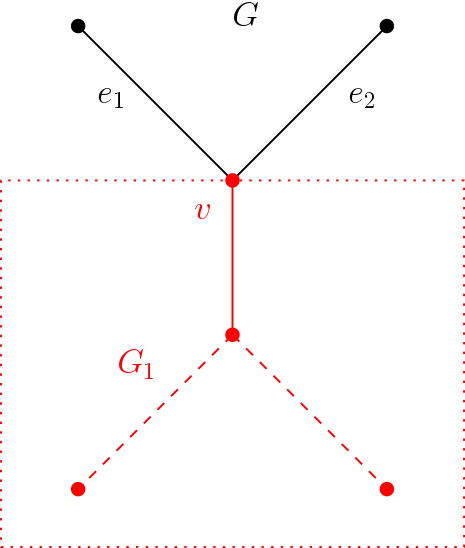}
\caption{Illustration of the proof of Case 2.}\label{gfcn4}
\end{figure}

By the induction hypothesis, two parallel flows $\mathcal{P}_1, \mathcal{P}_2$ cover $G_1$. Without loss of generality, we assume that there is a nontrivial path $P_1\in \mathcal{P}_1$ that contains $v$ as an end point. Then we can first extend $P_1$ by attaching $e_1$ to form $\mathcal{P}_1'$.

Suppose $\mathcal{P}_2$ does not contain any nontrivial path that ends at $v$. We add the path $e_1\cup e_2$ to $\mathcal{P}_2$ to form $\mathcal{P}_2'$. Clearly, both $\mathcal{P}_1'$ and $\mathcal{P}_2'$ are parallel flows and they cover $G$.

On the other hand, if $\mathcal{P}_2$ contains a path $P_2$ that also ends at $v$. Then we extend $P_2$ by attaching $e_2$ to form $\mathcal{P}_2'$. Again, both $\mathcal{P}_1'$ and $\mathcal{P}_2'$ are parallel flows and they cover $G$.

\item $d_{\max}$ is odd.
In this case, we apply induction to $d_{\max}$. The base case $d_{\max}=3$ is proven in \ref{casedmax_3} and we now consider $d_{\max}\geq 5$. We first claim that there is a parallel flow $\mathcal{P}$ such that for each $v\in V$, there is a path $P\in \mathcal{P}$ containing $v$. To construct such a parallel flow, one can first include any non-extendable path $P$. Then for each neighbor $v$ of any path already in $\mathcal{P}$ with degree $d_{\max}$, we include in $\mathcal{P}$ a non-extendable path parallel to $\mathcal{P}$. This procedure can be repeated until each $v$ with degree $d_{\max}$ is contained in a path in $\mathcal{P}$.

We construct a (possibly disconnected) new graph $G_1$ by removing the edges in $\mathcal{P}$ from $G$ and then isolated vertices (see \figref{gfcn5}). Each component of $G_1$ has maximal degree $d_{\max}-2$. Moreover, as we assume $d_{\max}\geq 5$, the vertices with degree $1$ in $G_1$ are of degree $1$ in $G$ as well. Therefore, by the induction hypothesis, we are able to find a cover of $G_1$ with $\floor{(d_{\max}-1)/2}$ parallel flows. Taking union with $\mathcal{P}$, we obtain a cover of $G$ with $\floor{(d_{\max}-1)/2}+1=\floor{(d_{\max}+1)/2}$ parallel flows.
\begin{figure}[!htb]
\centering
\includegraphics[scale=0.35]{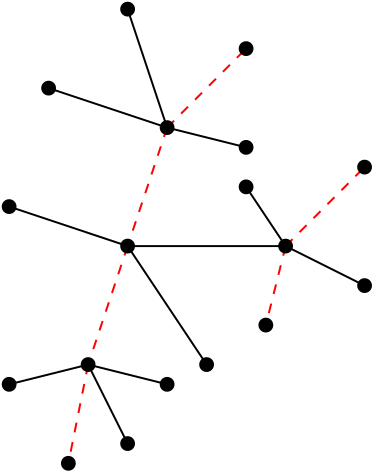}
\caption{This is an example with $d_{\max}=5$ of Case 3. The dashed red paths form the parallel flow $\mathcal{P}$. When we remove $\mathcal{P}$, we obtain $G_1$, which is a union of $3$ trees each with maximal degree $3$.}\label{gfcn5}
\end{figure}
\end{enumerate}	
\end{IEEEproof}	

Finally, we prove the general case. Given $G$, we construct a tree $G_1$ containing $G$ as follows: For each $v$ with degree $d_G(v)$ satisfying $1< d_G(v) < d_{\max}$, we add $d_{\max}-d_G(v)$ edges to $v$ (and new vertices) so that the degree of $v$ in $G_1$ is $d_{\max}$. Let $V_1'$ be the vertices of $G_1$ with degree $1$. Then, the condition in Lemma~\ref{lem:ivsv}\ref{it:sgia} holds true for $V_1$ and $V_1'$. From \cref{lem:ivsv}\ref{it:sgia} and \cref{lem:regular_tree}, $\mu(G,V')\leq \mu(G_1,V_1')=\floor{(d_{\max}+1)/2}$. The proof is now complete.

\section{Proof of Proposition~\ref{prop:sai}} \label{sec:pp}
    We prove the result for $S=A$ the graph adjacency matrix, and indicate at the end of the proof modification required for other choices of $S$.

	Suppose the polynomial for the GCN convolution layer is $p(x)= \sum_{1\leq i\leq m}a_{n_i}x^{n_i}$ such that $0\leq n_i < n_{i+1}$ and $a_{n_i}\neq 0, 1\leq i\leq m$. 
	
	To construct the corresponding GFCN model, we concatenate a convolution layer followed by a fusion layer with sum as the fusion function, repeated $n_m$ times. For each convolution layer, same padding is used and the same $3\times 1$ filter is applied to every parallel flow. In addition, the input layer has a direct skip layer connection to the $(n_m-n_i+1)$-th convolution layer, for each $i=1,\ldots,m$, where the $(n_m+1)$-th convolution layer is understood to be the output layer. 
	
\begin{figure}[!htb]
	\centering
	\includegraphics[scale=1]{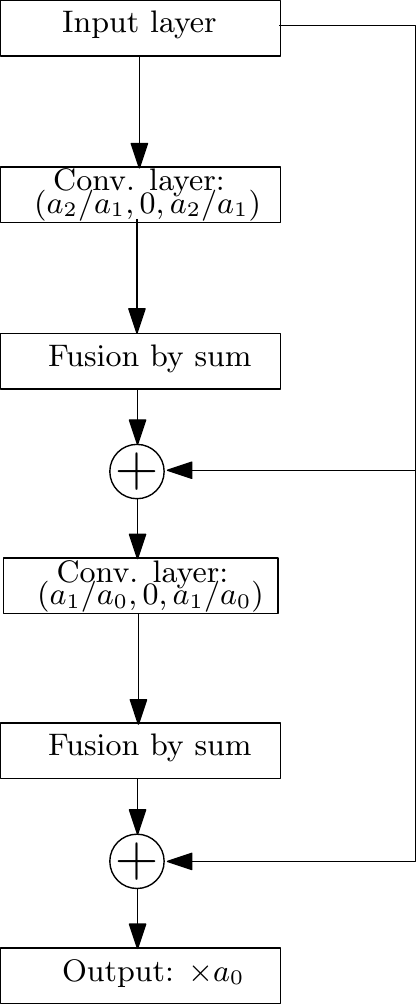}
	\caption{The diagram illustrates how to use GFCN to model the polynomial filter $p(A) = a_2A^2+a_1A+a_0$, where $A$ is the adjacency matrix.}\label{gfcn8}
\end{figure}

	To specify the $3\times 1$ filters, we consider different cases. If the layer does not have a direct connection from the input layer, the filter $(1,0,1)$ is used. If a layer is the $j$-th layer receiving a direct connection from the input layer that is not the output layer, then the $3\times 1$ filter $(a_{n_{m-j+1}}/a_{n_{m-j}},0,a_{n_{m-j+1}}/a_{n_{m-j}})$ is applied. If the output layer receives a direct connection from the input layer, then $n_1=0$ and we multiply $a_{n_1}$ at the output layer. An explicit example is shown in \figref{gfcn8}.
	
	It suffices to prove the following claim: if there is no skip layer connection, the output is the same the output of the convolution filter by $p(A)$ with $p$ being the monomial $a_{n_m}x^{n_m}$. This is because the skip layer connection corresponds to summation of monomials in a general $p(x)$. 
	
	For the claim, after each pair of convolution layer and fusion by taking summation, each node receive contribution from all of its neighbors in the original graph, weighted either by $1$ or $a_{n_{m-j+1}}/a_{n_{m-j}}$, exactly once (as we assume that no edge is repeated). This corresponds to applying $A$ once with the same weight. The concatenation of all these pairs of layers thus gives a monomial of degree $a_m$. The coefficient is the same as the product of all the weights, which is exactly $a_{n_m}$.   

A similar argument works for the other choices of $S$: $L$ the graph Laplacian, $\tilde{A}$ the normalized adjacency matrix and $\tilde{L}$ the normalized Laplacian. We summarize the changes to be made for the construction of the GFCN model in each case in Table~\ref{tab:5}, where ``Conv. filter (skip)" means the convolution filter to apply when receiving a direct connection from the input.

\begin{table}[!htb]
	\caption{GFCN models for different $S$.} \label{tab:5}
	\centering  
	\scalebox{1.3}{
		\begin{tabular}{|l|c|c|c|}  
			\hline
			 & \emph{Conv. filter (skip)}& \emph{Conv. filter} & \emph{Fusion function}  \\ 
			\hline  \hline 
			$A$ & $\frac{a_{n_{m-j+1}}}{a_{n_{m-j}}}(1,0,1)$ & $(1,0,1)$ & sum \\
			\hline
			$\tilde{A}$ & $\frac{a_{n_{m-j+1}}}{a_{n_{m-j}}}(1,0,1)$ & $(1,0,1)$ & average \\
			\hline
			$L$  & $\frac{a_{n_{m-j+1}}}{a_{n_{m-j}}}(-1,1,-1)$ & $(-1,1,-1)$ & sum \\
			\hline
			$\tilde{L}$& $\frac{a_{n_{m-j+1}}}{a_{n_{m-j}}}(-1,1,-1)$ & $(-1,1,-1)$ & average \\
			\hline
	\end{tabular}}
\end{table}

\bibliographystyle{IEEEtran}
\bibliography{IEEEabrv,StringDefinitions,refs}

\end{document}